\documentclass[conference]{IEEEtran}
\usepackage{times}

\IEEEoverridecommandlockouts

\usepackage[numbers]{natbib}
\usepackage{multicol}
\usepackage{multirow}
\usepackage[bookmarks=true]{hyperref}
\usepackage{float}
\usepackage{booktabs}
\usepackage{tabularx}
\usepackage[caption=false]{subfig}
\usepackage{amsmath}
\usepackage{listings}

\usepackage[]{graphicx}
\graphicspath{figures}

\newcommand{\pred}[1]{{\small\texttt{#1}}}
\newcommand{\var}[1]{{\textlangle\ignorespacesafterend #1\textrangle}}

\usepackage[dvipsnames]{xcolor}
\usepackage{FiraMono}

\newcolumntype{P}[1]{>{\centering\arraybackslash}p{#1}}
\newcolumntype{M}[1]{>{\centering\arraybackslash}m{#1}}

\usepackage{listings}
\begin{document}

\title{\LARGE \bf Bootstrapping Object-Level Planning with Large Language Models}


\author{David Paulius$^{1*}$, Alejandro Agostini$^{2}$, Benedict Quartey$^{1}$, George Konidaris$^{1}$\\
\thanks{$^{1}$Department of Computer Science, Brown University, USA}
\thanks{$^{2}$Department of Computer Science, University of Innsbruck, Austria}
\thanks{$^{*}$Corresponding Author (\textit{Email}:~\texttt{dpaulius@cs.brown.edu})}
}


\maketitle
\thispagestyle{empty}
\pagestyle{empty}

\renewcommand*{\thefootnote}{\fnsymbol{footnote}}


\begin{abstract}
    We introduce a new method that extracts knowledge from a large language model (LLM) to produce \textit{object-level plans}, which describe high-level changes to object state, and uses them to bootstrap task and motion planning (TAMP).
    Existing work uses LLMs to directly output task plans or generate goals in representations like PDDL. 
    However, these methods fall short because they rely on the LLM to do the actual planning or output a hard-to-satisfy goal.
    Our approach instead extracts knowledge from an LLM in the form of plan schemas as an object-level representation called functional object-oriented networks (FOON), from which we automatically generate PDDL subgoals.    
    Our method markedly outperforms alternative planning strategies in completing several pick-and-place tasks in simulation.\footnote[2]{Project Website:~\url{https://davidpaulius.github.io/olp_llm/}}
\end{abstract}

\renewcommand*{\thefootnote}{\arabic{footnote}}

\setcounter{footnote}{2}


\IEEEpeerreviewmaketitle

\section{Introduction}
	
The advent of \textit{large language models} (LLMs) has led to a plethora of work that exploits their capabilities for a variety of tasks, including planning for robotics~\citep{ahn2022saycan,driess2023palme} and embodied agents~\citep{huang2022language,raman2024cape}.
These approaches use LLMs as either a planner~\citep{ahn2022saycan,singh2023progprompt,driess2023palme}, or a goal generator~\citep{liu2023llm,xie2023translating,liu2024delta,kumar2024owltamp}.
As a task planner, an LLM is informed of the task and scene and directly outputs a complete plan, thus forgoing automated planning with off-the-shelf planners~\cite{ghallab_nau_traverso_2016}.
Plan actions generated by an LLM are then grounded to action policies or primitives.
As a task goal generator, an LLM generates planning definitions in the form of representations like PDDL~\citep{mcdermott1998pddl} (short for \textit{Planning Domain Definition Language}); this type of approach is often associated with task and motion planning (TAMP)~\citep{garrett2021integrated}.

However, existing work in these categories fails to handle complex, goal-oriented tasks in several key aspects.
On the one hand, positing the LLM as a task planner deprives such methods of guarantees promised by classical planning (viz. optimality and completeness).
Recent work has also called to question whether LLMs can effectively plan~\cite{valmeekam2023llm}.
On the other hand, using the LLM as a task description generator will fail to generate plan specifications that are guaranteed to work due to the LLM's lack of embodiment.
For instance, it may be difficult for the LLM to generate accurate PDDL definitions simply from a language description of the robot's environment.

It is natural to exploit language models for planning as they contain useful domain knowledge and often output useful steps.
Similarly, they are useful as goal generators because one can still rely on off-the-shelf planners.
This work uses an LLM to generate \textit{partial goal schemas at the object level}, which can then form PDDL subgoals.
Such an approach inherits the desirable commonsense planning knowledge of the LLM while supporting sound and complete task-level planning.
The object (as opposed to task) level is the level at which natural language is most appropriate and at which most knowledge is captured and expressed~\cite{kroemer2021review,paulius2022object}. 
While task-level planning focuses on action or motion constraints for execution, \textit{object-level planning} focuses on object interactions without committing to \textit{how} these effects will be resolved until runtime.



\begin{figure}
    \centering
    \includegraphics[width=0.8\columnwidth,trim={1cm 0cm 0cm 0cm},clip]{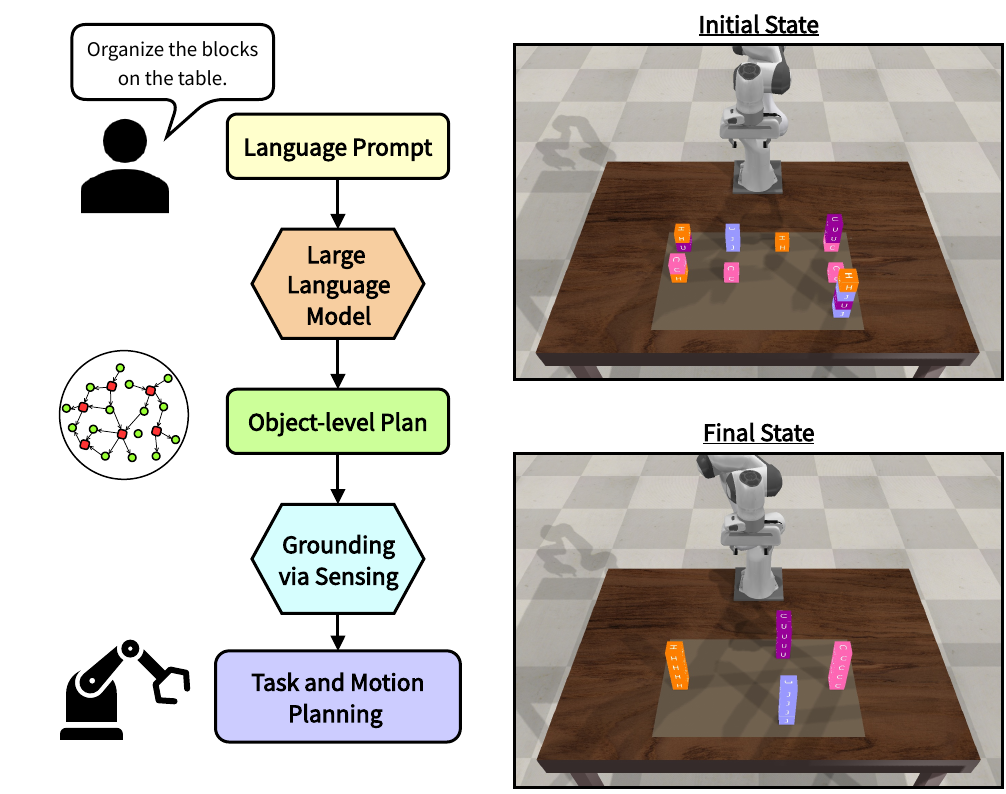}
    \vspace{-0.2cm}
    \caption{Our approach prompts an LLM for object-level information with which we construct an object-level plan (as a FOON). This plan schema \textit{bootstraps} task- and motion-level planning (TAMP) via PDDL subgoals.}
    \vspace{-0.6cm}
\label{fig:overview}
\end{figure}

We propose a modular approach that distills domain knowledge from an LLM to generate \textit{object-level plans}~\citep{paulius2022object}, which then bootstrap hierarchical planning.
We situate object-level planning as an interface between human language and TAMP and exploit an object-level representation (OLR) called the \textit{functional object-oriented network} (FOON)~\citep{paulius2016functional}. 
Recent work has shown how object-level knowledge in FOON can automatically generate PDDL subgoals~\cite{paulius2023longhorizon}; however, this assumes that partial plan specifications already exist as a FOON.
We exploit the capabilities of LLMs for object-level planning, overcoming the inability of LLMs to directly output feasible task plans while exploiting the higher, object-level nature of LLM output and language as a whole.


\begin{figure*}[t]
    \centering
    \includegraphics[width=0.80\textwidth]{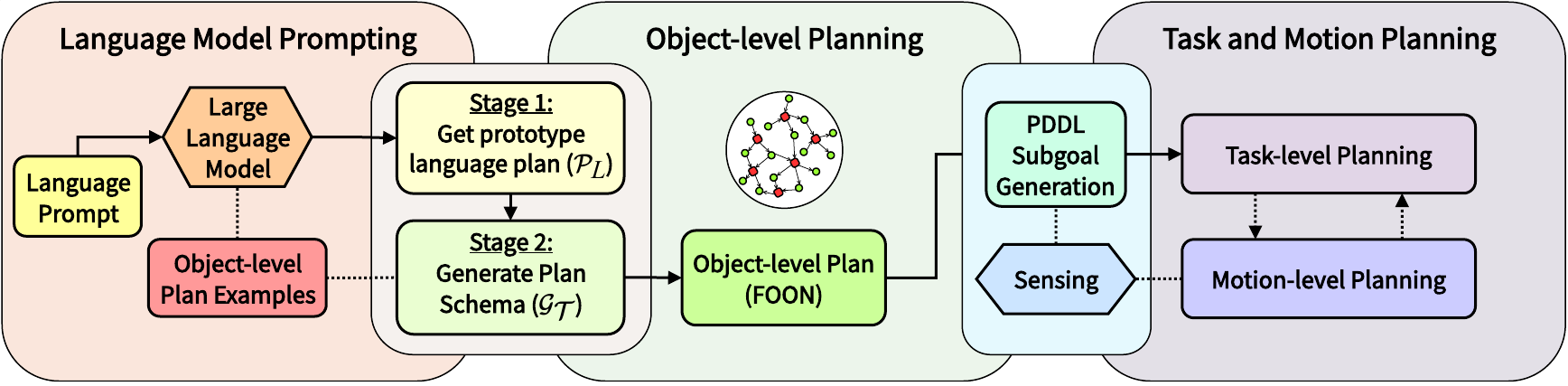}
    \vspace{-0.1cm}
    \caption{Our approach interfaces with a language model to generate object-level plans (as FOON graphs) for bootstrapping task and motion planning. We generate task-level subgoals as PDDL subgoals by grounding object-level subgoals to the robot's environment; with these task-level definitions, task planning to obtains task plan segments per object-level action, which are executed using motion-level planning, improving prior work~\cite{paulius2023longhorizon}.}
    \vspace{-0.3cm}
    \label{fig:my_label}
\end{figure*}

The contributions of our work are as follows: first, we introduce a modular planning approach (Figure~\ref{fig:overview}) that interfaces with an LLM to generate natural language instructions, from which we transform into an OLR (e.g., FOON) for hierarchical planning. Second, we show how object-level information can be distilled directly from an LLM and then used to generate planning definitions as PDDL, 
improving the feasibility of generated plans.
Finally, we showcase markedly better performance than alternative LLM-based methods.


\section{Background}
\label{sec:background}

\textbf{Large Language Models:}
\label{sec:background-llm}
A large language model (LLM) is a complex neural network model trained via self-supervised learning and self-attention~\citep{vaswani2017attention}.
LLMs have shown remarkable performance in natural language processing (NLP) and text generation tasks.
Variants such as GPT~\citep{brown2020language} and LLaMA~\citep{touvron2023llama} are trained on large corpora of text collected from the Internet and fine-tuned using RLHF (reinforcement learning from human feedback). 
For this reason, an LLM can be thought of as a ``compressed" representation of domain knowledge from the web~\citep{chiang2023chatgpt}, which is why we aim to exploit these models to inform planning.
This work uses OpenAI's Chat-GPT~\citep{openai_gpt4o}.

\textbf{Task and Motion Planning:}
The aim of task and motion planning (TAMP) is to integrate higher-level symbolic \textit{task planning} with lower-level \textit{motion planning} to enable robots to solve complex long-horizon tasks~\citep{garrett2021integrated}.
At the lower level, \textit{motion planning} finds collision-free robot motion or trajectories that are typically used to achieve a task.
However, typical robot tasks are too complex for motion-level planning alone.
For this reason, \textit{task planning} is necessary as an added layer to reason over an abstraction of the robot's actions and environment.
Task planning assumes a state description $\mathcal{S}$ using logical predicates, which are \pred{true} or \pred{false} depending on whether or not the robot observes them.
Starting from an initial state $s_0 \in \mathcal{S}$, task planning finds an action sequence $a \in \mathcal{A}$ that achieves a goal $g$ as a task plan $\mathcal{P} = \{a_1, ..., a_n\}$~\cite{ghallab_nau_traverso_2016}.
An action $a$ refers to a robot-executable skill or policy; our work assumes access to a repertoire of skills, which we denote by $\mathcal{A}$, which are defined as planning operators in PDDL~\citep{mcdermott1998pddl}.
Finally, given the task plan $\mathcal{P}$, motion planning finds collision-free trajectories that reproduce the intended effects of each action; our work uses OMPL~\cite{sucan2012ompl} for motion planning.


\subsection{Related Work}
\label{sec:related}


\textbf{Language Models for Planning}:
Many researchers have explored the use of language models for robotics applications, having been inspired by their remarkable performance in language-related tasks.
Prior works have investigated the planning capabilities of LLMs~\cite{silver2023generalized,valmeekam2023llm}.
Other works supplement task planning with language models~\citep{liu2023llm,chen2024autotamp,singh2023progprompt,singh2024twostep,liu2024delta,han2024interpret}.
LLM+P~\citep{liu2023llm} generates PDDL problem file via LLM prompting.
Much like our work, existing works use LLMs as an informer of subgoals for classical planning~\citep{singh2024twostep,liu2024delta,han2024interpret,kumar2024owltamp}.
In particular, DELTA~\citep{liu2024delta} resembles our method in that it decomposes a task into a series of PDDL subgoal definitions directly output by an LLM.
Our approach uses an LLM at the object level and not task level (i.e., PDDL).
Recent work also iteratively prompts an LLM for FOON generation~\cite{sakib2024consolidating}.
Similar to DELTA, they do not focus on generating nor executing physically valid plans.

\textbf{Language Models as Planners}:
Several works treat language models as robotic task planners.
SayCan~\citep{ahn2022saycan} combines a language model and affordance detectors for driving robotic execution given a task prompt.
PaLM-E~\citep{driess2023palme} is an embodied language model that directly incorporates continuous observations (like images, state estimates, or other sensor modalities) into the language embedding space.
These works have shown that language models are capable of performing some degree of embodied reasoning.
However, one major drawback of these works is that they require a large amount of engineering effort, particularly to enable them to operate in novel environments and solve long-horizon tasks.
Previous works also exploited the reasoning capabilities of an LLM to solve a wide range of tasks both in simulation~\citep{huang2022language,gramopadhye2023generating,raman2024cape} and with a real robot~\citep{raman2024cape}.



\section{Object-level Planning with Language Models}
\label{sec:llm}
There exists a disconnect between language and the task level, which makes TAMP unsuitable for generalization across tasks and settings.
Yet, existing works use LLMs either as planners or task description generators for task execution; these approaches fall short because of the inability of LLMs to correctly reason about task- and motion-level constraints.
It is impractical to provide the entire context of a task setting to an LLM and expect it to handle all the reasoning about a robot's embodiment (e.g., where objects are located, in what poses they are, what type of gripper the robot has, etc.) in order to generate adequate planning definitions or feasible task plans.

Instead, the strength of language models lies in their ability to provide approximate subgoals that are useful to decision making at both task and motion levels.
This is because language models can express task-relevant knowledge in a generic yet informative way.
Imagine your typical cooking recipe, for instance: a recipe provides a sketch of object interactions agnostic to the state of the reader's kitchen or the recipe writer's kitchen.
It also does not provide details on how actions should be executed (e.g., which hand should be used, how should an object be grasped, etc.).
What a recipe expressed in natural language may provide, however, is an idea of the types of actions and inter-object interactions necessary to complete a task.
Rather, the exact details of task- and motion-level execution are resolved at run time. 

For these reasons, we adopt an \textit{object-level planning} approach to bootstrap task and motion planning~\cite{paulius2023longhorizon}.
We generate object-level plan sketches, which provide task-level subgoals that naturally interface language and decision making, using an LLM.
Briefly, given a language command to a robot, our approach (Figure~\ref{fig:my_label}) uses an LLM to generate a sequence of natural language instructions, which is then transformed into an \textit{object-level plan} (OLP) represented as a FOON.
It is then through task planning where properties relevant to the robot (e.g., robot's end-effector and object poses) are used to find a task-aware plan which is then executed via TAMP.
Task-level planning is achieved by transforming each OLP action into PDDL definitions to find task plan segments.

\subsection{Object-level Planning}
\label{sec:background-foon}
We adopt another layer of planning above TAMP called object-level planning, which considers changes to object state~\citep{paulius2022object}.
We use an object-level representation in the form of a knowledge graph called the \textit{functional object-oriented network} (FOON)~\citep{paulius2016functional,paulius2019survey}.
Formally, a FOON $\mathcal{G} = \{\mathcal{O}, \mathcal{M}, \mathcal{E}\}$ is a bipartite graph with object nodes ($o \in \mathcal{O}$) and motion nodes ($m \in \mathcal{M}$) connected via directed edges ($e \in \mathcal{E}$), which reflect the change of an object's state as it is manipulated via a corresponding action.
An object $o = (o_{t}, o_{s}, o_{\mathcal{I}})$ is defined as a tuple with the following attributes: its object type or name ($o_{t}$), its state ($o_{s}$), and, if applicable, its object composition ($o_{\mathcal{I}} = \{o_{t_1}, o_{t_2}, ..., o_{t_n}\}$, where $n = |\mathcal{I}|$).
A motion node $m = (m_t)$ is defined by an action verb or type ($m_t$).
A FOON describes object-state transitions via \textit{functional units} ($\mathcal{FU} = \{\mathcal{O}_{in}, \mathcal{O}_{out}, \tilde{m}\}$) at a level close to human language.
A functional unit defines preconditions and effects of executing an action ($\tilde{m}$), where a set of input nodes ($\mathcal{O}_{in}$) are required to produce a new set of output object nodes ($\mathcal{O}_{out}$).
We illustrate an example of a functional unit in Figure~\ref{fig:llm-to-olp}, which describes an action for a block stacking task.
We argue that foundation models naturally interface with object-level representations due to their similarity to human language, which in turn allows them to interface with tools that combine vision and language~\citep{brown2020language,zhang24survey}.





\subsection{LLM Prompting to Object-level Plan}
\label{sec:llm-iter}

\begin{figure}[t]
    \centering
    \includegraphics[width=\columnwidth]{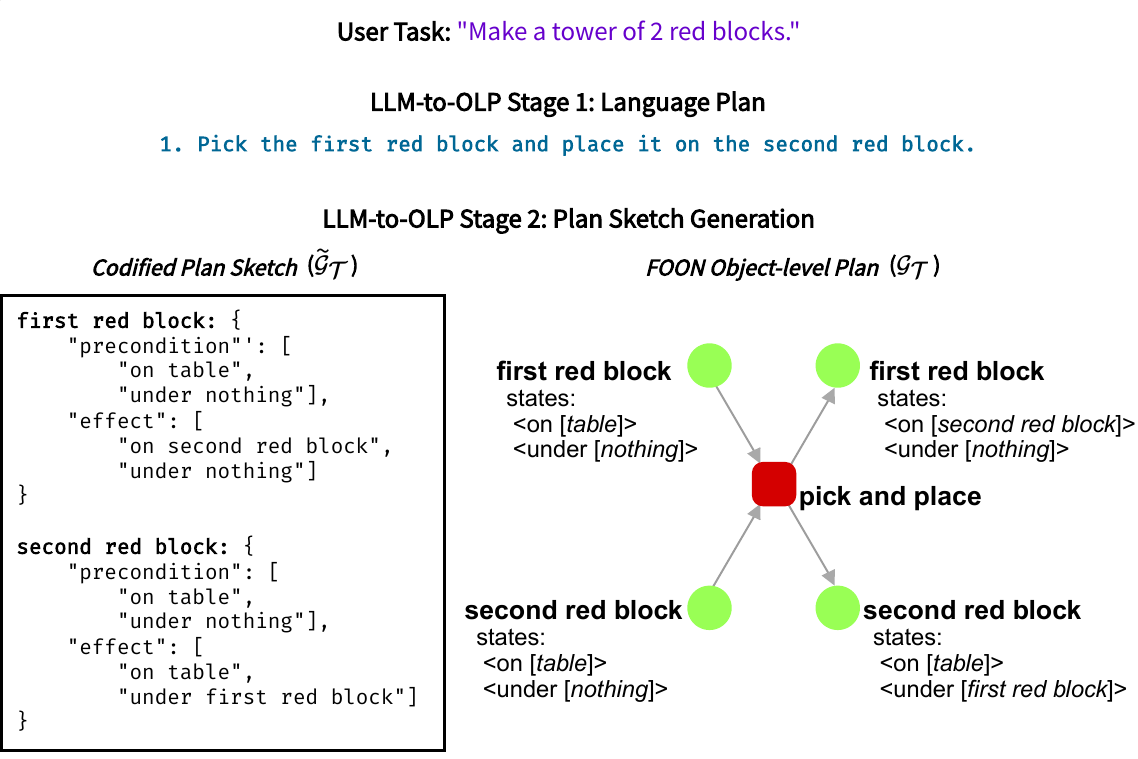}
    \vspace{-0.5cm}
    \caption{Illustration of how a user task specified in natural language is transformed into an object-level plan (OLP) as a FOON via LLM prompting.}
    \label{fig:llm-to-olp}
    \vspace{-0.3cm}
\end{figure}

Our goal is to extract attributes for object-state transitions before and after each action is performed (i.e., preconditions and effects) to construct an OLP describing task subgoals.
We instantiate object-level planning with FOONs.
Our approach constructs a FOON $\mathcal{G}_{\mathcal{T}}$, where $\mathcal{T}$ is a task given in natural language (such as \textit{``Make a tower of two red blocks"}---see Figure~\ref{fig:llm-to-olp}) via a two-stage process.
In addition to the task instruction $\mathcal{T}$, we include a language description of objects in the scene (from which the LLM will determine those that are relevant to the task) as well as a set of example object-level plans (as FOONs) 
$\mathcal{X}_{\mathcal{G}}=\{\mathcal{G}_1, \mathcal{G}_2,...,\mathcal{G}_n\}$ for reference.

The first stage prompts an LLM for a plan sketch comprised of natural language instructions denoted by $\mathcal{P}_{L} = \{\xi_1, \xi_2, ..., \xi_n\}$, where $\xi_i$ refers to an instruction as text.
As an example in Figure~\ref{fig:llm-to-olp}, given a task and available objects (without \textit{any} context about their present configuration), we expect text instructions $\mathcal{P}_{L}$ that solely mention red blocks for the task \textit{``Make a tower of 2 red blocks.''}
During this step, we transform the top-$k$ most similar FOONs in $\mathcal{X}_{\mathcal{G}}$ into example plan sketches, from which the LLM selects the one closest to the new task to use as reference (denoted by $\hat{\mathcal{G}}$).
We identify the top $k$ candidates using cosine similarity between text embeddings of the task prompt $\mathcal{T}$ and the set of instructions for a given reference $\hat{\mathcal{G}} \in \mathcal{X}_{\mathcal{G}}$.
An example sketch may describe how three generic blocks (regardless of type) can be stacked into a tower.
In the second stage, the LLM must reason about each instruction $\xi_i \in \mathcal{P}_{L}$ to generate an OLP for the novel task. 
We prompt the LLM to reason about state changes of task-relevant objects, specifically geometric relations for task-level planning (Section~\ref{sec:foon-to-pddl}).
In the previous example, we expect output with state descriptions such as \textit{``first red block on second red block,"} \textit{``first red block under nothing,"} \textit{``second red block under first red block,"} and \textit{``second red block on table"} (Figure~\ref{fig:llm-to-olp}).
We assist the LLM by providing $\hat{\mathcal{G}}$ in the prompt, with which it must generate a new FOON $\mathcal{G}_\mathcal{T}$ for the novel task.
Inspired by previous work on code writing for robots~\cite{liang2023code}, we codify $\hat{\mathcal{G}}$ as a JSON.
The LLM then outputs a codified OLP $\tilde{\mathcal{G}}_\mathcal{T}$, which captures each instruction $\xi_i \in \mathcal{P}_{L}$. 
Finally, each action in $\tilde{\mathcal{G}}_\mathcal{T}$ forms a functional unit $\mathcal{FU}_i \in \mathcal{G}_\mathcal{T}$.

\section{Bridging to Task and Motion Planning}
\label{sec:hierarchy}
We generate a plan schema $\mathcal{G}_{\mathcal{T}}$, with which we can solve a task $\mathcal{T}$ given in natural language.
However, this schema is too abstract to be executed in its present form, and it must be grounded to the robot's embodiment and environment~\citep{paulius2023longhorizon}.
Therefore, we use $\mathcal{G}_{\mathcal{T}}$ to \textit{bootstrap} TAMP via PDDL subgoals.
This is done through a hierarchical approach that automatically transforms each object-level action in $\mathcal{G}_{\mathcal{T}}$ into PDDL problem definitions and searches for a robot-executable plan given a predefined set of robot skills or operators~\citep{paulius2023longhorizon}.

\subsection{Object-Level to Task-Level Planning}
\label{sec:foon-to-pddl}
The aim of task-level planning is to find a robot-executable \textit{task plan} $\mathcal{P}_\mu$ that solves task $\mathcal{T}$.
A task plan is composed of a sequence of smaller plan segments for each functional unit, i.e., $\mathcal{P}_\mu = \{\tilde{\mathcal{P}}_{\mu_1}, \tilde{\mathcal{P}}_{\mu_2}, ..., \tilde{\mathcal{P}}_{\mu_n}\}$, where
$\tilde{\mathcal{P}}_{\mu_i} = \{a_{\mu_1}, a_{\mu_2}, ..., a_{\mu_m}\}$ denotes a plan segment achieving the subgoals described by a functional unit $\mathcal{FU}_i$ and $a_{\mu_i}$ refers to the $i$-th step corresponding to a parameterized skill in $\mathcal{A}$.

\label{sec:foon-to-pddl-pred}
PDDL solvers require two components: a \textit{domain definition} and a \textit{problem definition}~\cite{mcdermott1998pddl}.
A domain definition provides details on what actions can be taken by a robot as well as possible object types, while a problem definition captures the initial state of the robot and its environment ($s_\mu$) as well as the target goal state ($g_\mu$) as logical predicates.
We assume a predefined domain definition with planning operators corresponding to a repertoire of parameterized, robot-executable skills $\mathcal{A}$.
For the generation of problem definitions, both $s_\mu$ and $g_\mu$ are adapted from a functional unit $\mathcal{FU} \in \mathcal{G}_\mathcal{T}$: predicates are constructed based on the object-state pairs in $\mathcal{FU}$~\cite{paulius2023longhorizon}. 
In other words, transforming a FOON into PDDL requires mapping attributes of each object $o$ to predicates (where $o \in \mathcal{O}_{in} \cup \mathcal{O}_{out}$).

\textbf{Object-centered Predicates:}
We use object-centered predicates~\citep{agostini2020manipulation,paulius2023longhorizon} that describe constraints for collision-free motion.
They are written as \pred{(\var{rel} \var{?obj\_1} \var{?obj\_2})}, where \pred{\var{rel}} refers to a geometric relation using the spatial adpositions \textit{in}, \textit{on}, or \textit{under}, while \pred{\var{?obj\_1}} and \pred{\var{?obj\_2}} refer to objects described by a given predicate. 
These relations are described from the reference frame of each object, which permits propagating motion constraints at task planning for the generation of feasible plans~\citep{agostini2023unified}. For example, the predicate \pred{({on} {block\_1} {block\_2})} denotes that \pred{block\_2} lies on top of \pred{block\_1}.
We also use a virtual object \textit{air} to describe free space in or on objects, which is important for collision-free picking, i.e., \pred{({on} {block\_1} {air})}---nothing is on a block, which makes it free for grasping).

\begin{figure}[t]
    \centering
    \includegraphics[width=0.9\columnwidth]{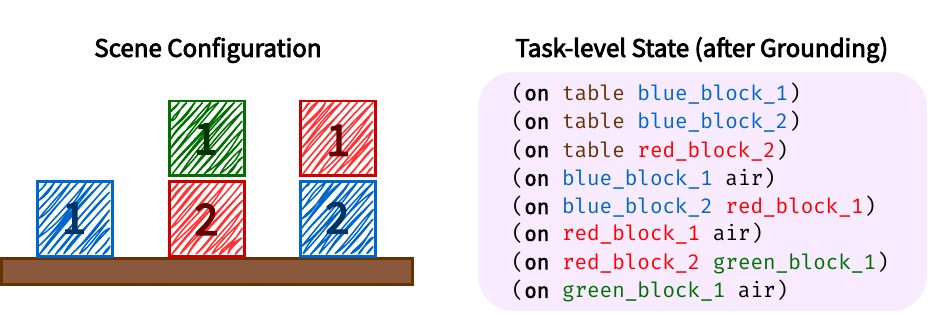}
    \vspace{-0.2cm}
    \caption{Example of task-level grounding for an object-level plan (Figure~\ref{fig:llm-to-olp}), which is compatible by design with the planning operators in Figure~\ref{fig:micro-PO}.}
    \label{fig:olp-to-tlp}
    \vspace{-0.4cm}
\end{figure}

\textbf{Grounding:}
Each subgoal in an OLP (i.e., functional unit in FOON) must be grounded to the robot's environment for effective task-level planning.
For starters, object-level aliases must be linked to object references at the task level.
This can be likened to how we as humans use recipes: a recipe refers to ingredients with words, but we must resolve their references to object instances around us when completing a recipe.
This work assumes that there exists an exact mapping of objects described in an OLP to those existing in the environment, and we prompt the LLM to map each alias to an instance.
For example, if we have two red blocks as objects in an OLP (Figure~\ref{fig:llm-to-olp}), an LLM will map them to object instances \pred{red\_block\_1} and \pred{red\_block\_2} (Figure~\ref{fig:olp-to-tlp}).
Once completed, we obtain a mapping of object-state pairs to task-level predicates: we use object poses (both position and orientation) and bounding boxes to derive object-centered predicates for each object $o$ in $\mathcal{G}_{\mathcal{T}}$ using the mechanism from previous work~\citep{agostini2023unified}.

\subsection{Task-Level to Motion-Level Planning}
\label{sec:foon-to-pddl-sense}
With each plan segment $\tilde{\mathcal{P}_{\mu}} \in \mathcal{P}_{\mu}$, a robot can then execute a sequence of actions that resolve object-level subgoals.
We use motion-level planning to find collision-free robot movements that will achieve the effects of a robot's skills.
This work considers picking and placing actions (Figure~\ref{fig:micro-PO}).
For the \textit{pick} action (Figure~\ref{fig:micro-po-a}), we generate a trajectory that moves the robot's end-effector from its initial position to a target object, while the \textit{place} action (Figure~\ref{fig:micro-po-b}) moves the robot's end-effector grasping an object from its initial pose to a position above a target surface or object. The initial and final poses of the hand for these actions can be obtained directly from object-centered hand-object relations encoded in the preconditions and effects of their corresponding planning operators using geometric rotation and translation transformations~\cite{agostini2023unified}.

\begin{figure}[t]
    \centering
    \vspace{-0.2cm}
    \subfloat[Pick Action]{
    \label{fig:micro-po-a}
    \texttt{\tiny
    \begin{tabular}[t]{l}
      	(\textbf{:action} pick \\
    	\indent~\textbf{:parameters} ( \\
            \indent~\indent~\textit{?obj} - object \\
            \indent~\indent~\textit{?surface} - object) \\
    	\indent~\textbf{:precondition} (and \\
            \indent~\indent~\textit{; collision-free constraints:} \\
    	\indent~\indent~(in hand air) (on \textit{?obj} air) \\
            \indent~\indent~\textit{; object is on a surface:} \\
    	\indent~\indent~(on \textit{?surface} \textit{?obj}) \\
    	\indent~\indent~(under \textit{?obj} \textit{?surface}) ) \\
    	\indent~\textbf{:effect} (and \\
            \indent~\indent~\textit{; hand contains target object:} \\
    	\indent~\indent~(in hand \textit{?obj}) (not (in hand air)) \\
            \indent~\indent~\textit{; object has been grasped:} \\
    	\indent~\indent~(on \textit{?obj} hand) \\
    	\indent~\indent~(under \textit{?obj} air) \\
    	\indent~\indent~(not (on \textit{?obj} air)) \\
            \indent~\indent~\textit{; nothing is on surface:} \\
    	\indent~\indent~(not (on \textit{?surface} \textit{?obj}))\\
    	\indent~\indent~(not (under \textit{?obj} \textit{?surface})) \\
    	\indent~\indent~(on \textit{?surface} air) ))\\
    \end{tabular}
    }
    }
    \subfloat[Place Action]{
    \label{fig:micro-po-b}
    \texttt{\tiny
    \begin{tabular}[t]{l}
      	(\textbf{:action} place \\
    	\indent~\textbf{:parameters} ( \\
            \indent~\indent~\textit{?obj} - object \\
                \indent~\indent~\textit{?surface} - object) \\
    	\indent~\textbf{:precondition} (and \\
            \indent~\indent~\textit{; collision-free constraints:} \\
    	\indent~\indent~(on \textit{?surface} air) \\
    	\indent~\indent~(under \textit{?obj} air)\\
            \indent~\indent~\textit{; hand contains object:} \\
    	\indent~\indent~(in hand \textit{?obj}) (on \textit{?obj} hand) ) \\
    	\indent~\textbf{:effect} (and \\
            \indent~\indent~\textit{; hand no longer contains object:} \\
            \indent~\indent~(in hand air)~(not (in hand \textit{?obj}))\\
            \indent~\indent~\textit{; object is on surface:} \\
    	\indent~\indent~(on \textit{?surface} \textit{?obj}) \\
    	\indent~\indent~(not (on \textit{?surface} air))\\     	                   \indent~\indent~(under \textit{?obj} \textit{?surface}) \\
    	\indent~\indent~(not (under \textit{?obj} air))\\
            \indent~\indent~\textit{; nothing is on object:} \\
            \indent~\indent~(not (on \textit{?obj} hand)) \\
    	\indent~\indent~(on \textit{?obj} air) ))\\
    \end{tabular}
    }
    }
    \caption{Planning operators for \textit{pick} and \textit{place} actions using object-centered predicates~\cite{agostini2020manipulation} and executable via motion-level planning (Section~\ref{sec:foon-to-pddl-sense}).
    }
    \label{fig:micro-PO}
    \vspace{-0.5cm}
\end{figure}

\begin{table*}[t]
    \centering
    \footnotesize
    \caption{Experimental results for several block stacking tasks across 10 trials per setting and block counts}
    \begin{tabular}{M{0.9cm}M{2cm}ccccc}
         \toprule
         \textbf{Task Setting} & \textbf{Planning Approach} & \textbf{\% Plan Complete}~$\uparrow$ & \textbf{\% Success}~$\uparrow$ & \textbf{Avg. Plan Time (s)}~$\downarrow$ & \textbf{Avg. Tokens}~$\downarrow$ & \textbf{Avg. Plan Length}~$\downarrow$ \\
         \midrule
         \multirow{4}{*}{\textit{Tower}} & OLP & $\mathbf{86.00\%}$ & $\mathbf{76.00\%}$ & $\mathbf{0.0043 \pm 0.0021}$ & $2406.38 \pm	335.0091$ & $10.2791 \pm 4.5687$ \\
         & LLM-Planner & $44.00\%$ & $34.00\%$ & $12.0486 \pm 6.3784$ & $\mathbf{744.94	\pm 120.9533}$ & $10.2791 \pm	4.5687$ \\
         & LLM+P & $18.00\%$ & $34.00\%$ & $0.0346 \pm	0.0312$ & $1656.42 \pm	170.2912$ &  $\mathbf{8.5556	\pm 3.9291}$ \\
         & DELTA & $\mathbf{86.00\%}$ & $60.00\%$ & $0.0067 \pm	0.01206$ & $4871.88	438.2610$ & $9.0233 \pm	4.7883$\\
         \midrule
         \multirow{4}{*}{\textit{Spelling}} & OLP & $\mathbf{80.00\%}$ & $\mathbf{62.00\%}$ & $0.02715 \pm	0.0828$ & $2588.66 \pm	379.1376$ & $\mathbf{8.45 \pm 	3.4932}$\\
         & LLM-Planner & $22.00\%$ & $16.00\%$ & $7.5734 \pm 	2.4650$ & $\mathbf{754.06 \pm	102.7516}$ & $9.7778	\pm 3.3529$  \\
         & LLM+P & $30.00\%$ & $46.00\%$ & $0.0268 \pm 
 0.0266$ & $1671.26 \pm	189.9115$ & $9.0435 \pm	4.0393$ \\
         & DELTA & $78.00\%$ & $50.00\%$ & $\mathbf{0.0075 \pm	0.0061}$ & $4836.72 \pm	475.0059$ & $10.5641 \pm	5.5998$\\
         \midrule
         \multirow{4}{*}{\textit{Organize}} & OLP & $\mathbf{81.43\%}$ & $\mathbf{77.14\%}$ & $\mathbf{0.0080 \pm	0.0053}$ & $3051.5 \pm 499.4818$ & $15.3684 \pm 7.1630$ \\
         & LLM-Planner & $35.71\%$ & $22.86\%$ & $24.1510 \pm 	15.9711$ & $\mathbf{885.0571 \pm 	126.3086}$ & $\mathbf{8.40 \pm 2.2361}$\\
         & LLM+P & $37.14\%$ & $37.14\%$ & $0.0538 \pm 0.1139$ & $1891.3286 \pm 	212.4083$ & $11.3077 \pm 4.1547$ \\
         & DELTA & $67.14\%$ & $54.29\%$ & $0.0139 \pm 	0.0312$ & $5329.9571 \pm 	470.9973$ & $13.8298 \pm 	6.4042$\\
         \bottomrule
    \end{tabular}
    \label{tab:results}
\end{table*}

\section{Evaluation}
\label{sec:result}
We evaluate the flexibility of our approach (denoted as \textbf{OLP} in Table~\ref{tab:results}) with alternative methods on several tasks in simulated experiments.
Our results show that an LLM cannot reliably produce PDDL definitions and is unable to reliably task plan due to its lack of spatial understanding; however, we can prompt an LLM for object-level subgoals compatible with our modular approach from previous work~\cite{paulius2023longhorizon}.

\subsection{Experimental Setup}
\label{sec:exp-setup}

We perform experiments in a simulated table-top environment in CoppeliaSim~\cite{coppeliaSim} with a Franka Emika Panda robot affixed to a table upon which blocks are randomly initialized.
Given a task specified in natural language, the robot must perform a sequence of \textit{pick} and \textit{place} actions (defined in Figure~\ref{fig:micro-PO}) fulfilling the task.  
We assume that the state of the environment is fully observable---object poses and bounding boxes are known via perception.
This information is used in motion-level planning to generate collision-free trajectories.
In addition to Chat-GPT\footnote{We tested \texttt{gpt-4}, \texttt{gpt-4o}, and \texttt{chatgpt-4o-latest}, but found \texttt{chatgpt-4o-latest} to produce the best plans, adhering to instructions.}~\citep{openai_gpt4o} as our LLM of choice, we use Fast Downward~\citep{helmert2006fast}, an off-the-shelf PDDL solver, for task-level planning in our method and baselines (discussed in Section~\ref{sec:baseline}). 
When planning with Fast Downward, we use the A* algorithm with the landmark cut (LMCUT) heuristic for plan optimality.
For motion-level planning, we use RRT-Connect~\cite{kuffner2000rrtconnect} as provided by OMPL~\cite{sucan2012ompl}.

\begin{figure}[t]
    \centering
    \includegraphics[width=\linewidth,trim={0.6cm 0cm 0cm 0cm},clip]{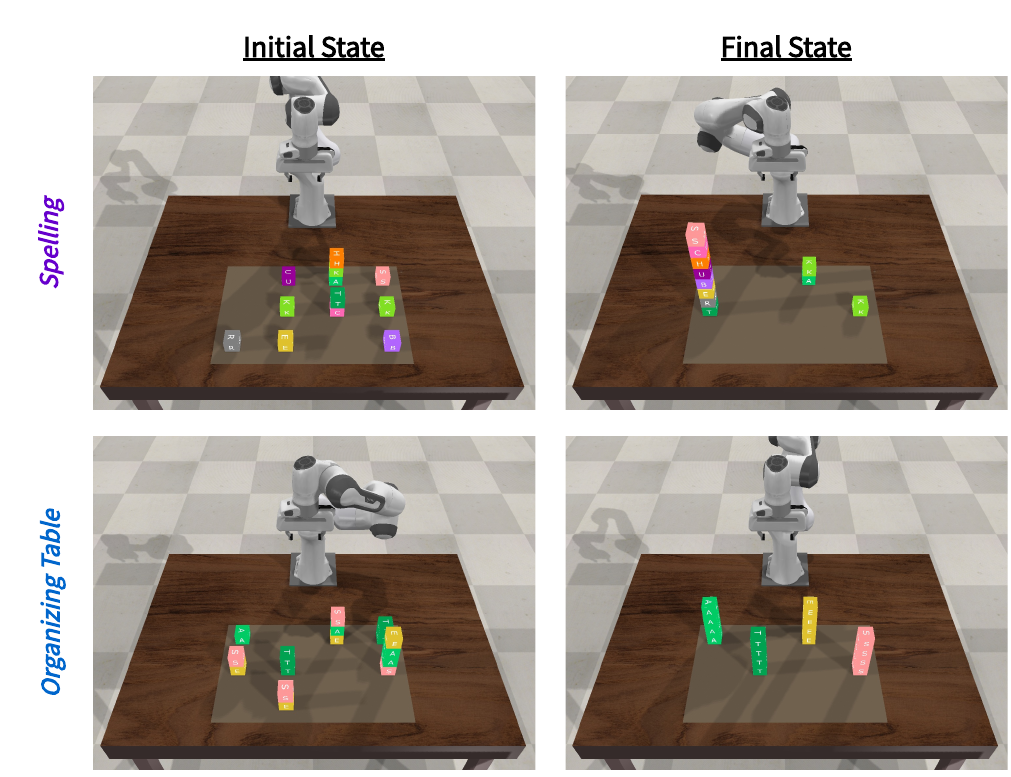}
    \vspace{-0.4cm}
    \caption{Example of initial and final states for the \textit{spelling} and \textit{organizing table} tasks. The \textit{tower building} task is akin to \textit{spelling} without ordering constraints.}
    \label{fig:setting}
    \vspace{-0.4cm}
\end{figure}

\textbf{Task Settings:}
We design scenarios in which the robot has to complete several tasks for three tasks of increasing difficulty: 1) tower building, 2) spelling, and 3) organizing a table (Figure~\ref{fig:setting}).
The \textit{tower building} task involves the robot assembling a tower of blocks of a given height $n$, where $3 \leq n \leq 7$, with $n+1$ blocks provided on the table.
The \textit{spelling} task also involves robot constructing a tower of blocks of some height $n$, but with the added constraint that the blocks correctly spell a given word of length $n$. 
This requires correct placement of lettered blocks, thus heavily depending on the LLM's ability to generate the correct sequence of pick and place actions.
Finally, the \textit{organizing} task involves a robot making piles of matching blocks: here, we initialize a scene of $3$ block types, each with $n$ block instances (where $2 \leq m \leq 4$).
This can be seen as a mix of the two prior tasks, where alike but varying numbers of blocks must be placed into piles.

\textbf{Metrics:} We report the following metrics: 1) \textit{plan completion}, which measures the percentage of all plans that were executed from start to finish \textit{regardless} of whether the task objective was achieved; 2) \textit{success}, which measures the percentage of successfully executed plans that achieve the task objective; 3) \textit{average plan computation time} (in seconds); 4) \textit{average number of tokens} for LLM prompting; and 5) \textit{average plan length} across all successful executions.

\subsection{Baseline Methods}
\label{sec:baseline}
We compare our OLP-based method to several baseline methods, for which we provide details below.
These baseline methods also rely upon Chat-GPT to either directly output a task plan or PDDL definitions, following the tracks of LLM-based planning work previously introduced.

\subsubsection{LLM-Planner}
This baseline serves as a proxy for methods that directly plan with LLMs~\citep{ahn2022saycan,driess2023palme}.
We directly provide a textual description of the state of the robot's environment (denoted by $\tilde{s}$) and the robot's executable skills ($\mathcal{A}$) as input and retrieve a task plan $\mathcal{P}_{\mu}$ as output.
We then parse each action to identify target objects and surfaces needed for pick and place actions while performing the necessary motion-level planning to successfully resolve each action.
This baseline approach evaluates the LLM's ability to reason about the robot's embodiment and produce a correct task plan.

\subsubsection{LLM+P}
LLM+P~\cite{liu2023llm} uses an LLM to generate a PDDL problem definition given a text description of a task planning domain and the initial state of the scene.
As input to the LLM, we provide a description of the robot's environment ($\tilde{s}$) and an example of a problem definition task, and we obtain a problem definition for task $\mathcal{T}$ as output. 
We then use this output with our domain definition of predefined skills to acquire a task plan using Fast Downward~\cite{helmert2006fast}, and this task plan is executed and resolved with motion-level planning.
This baseline approach evaluates the LLM's ability to accurately generate a PDDL problem file, compatible with a predefined set of skills, without explicitly performing object-level planning and reasoning.


\subsubsection{DELTA}
DELTA~\cite{liu2024delta} is a task planning method that auto-regressively prompts an LLM to derive PDDL domain and problem definitions.
Similar to our approach, a task $\mathcal{T}$ is broken down into subgoals, each of which is formulated as their own subgoal PDDL problem file.
We prompt the LLM with details about robot actions ($\mathcal{A}$) as well as the objects available to the robot, after which a domain file is generated.
The LLM is then provided with a state description $\tilde{s}$ and a task prompt $\mathcal{T}$ to generate a problem file that contains all goals (similar to the output of LLM+P~\cite{liu2023llm}).
This problem file is then broken down into subgoal problem files based on PDDL subgoals auto-regressively suggested by the LLM; this scopes the problem into subgoal actions that are akin to functional units.
We hypothesize that although this method will create simpler and smaller problem definitions, it heavily relies on the LLM's ability to generate syntactically and semantically correct definitions, which may not be as reliable as our method.

\subsection{Results and Discussion}
\begin{figure}[t]
    \centering
    \includegraphics[width=\linewidth]{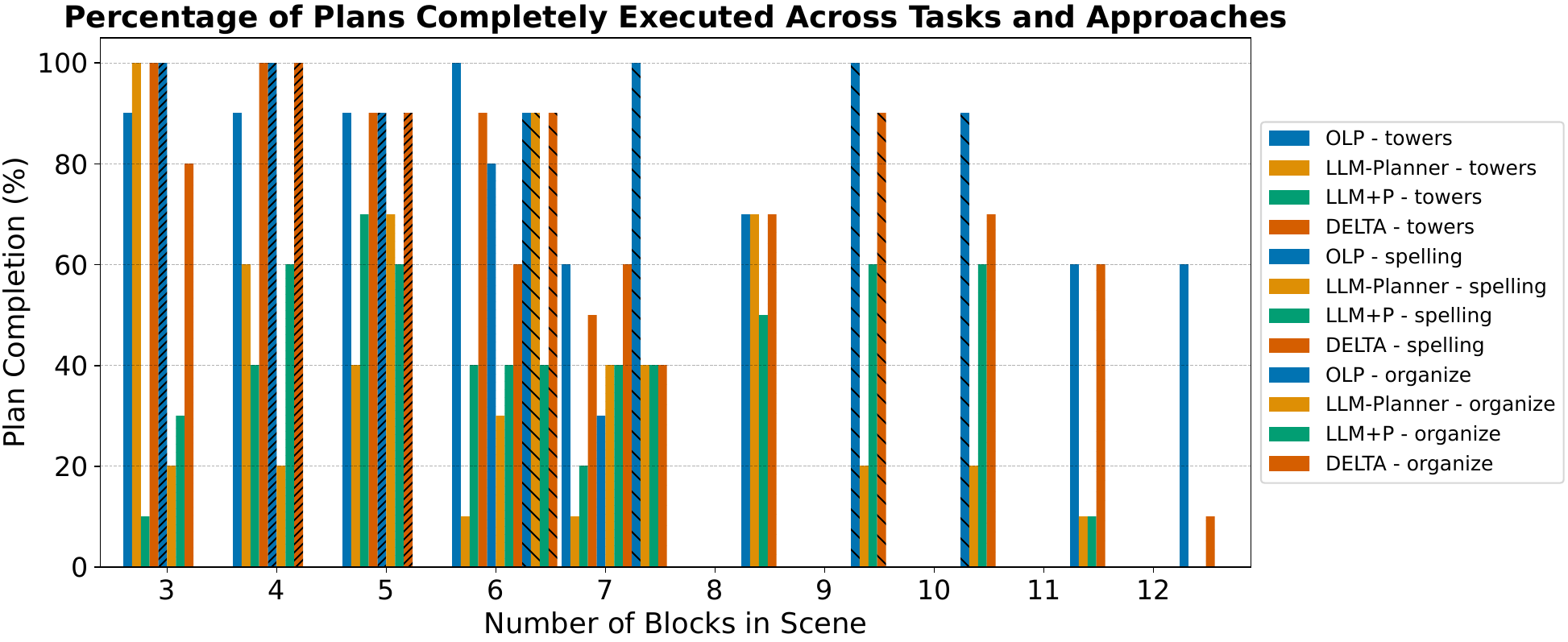}
    \vspace{-0.4cm}
    \caption{Graph showing percentage of plans completely executed using all approaches for different number of blocks across tasks (best viewed in colour).}
    \label{fig:enter-label}
    \vspace{-0.5cm}
\end{figure}

Our experimental results show that our OLP-based method performs better than baselines that either directly generate a task plan or PDDL files (Table~\ref{tab:results}).
Across all tasks and evaluated approaches, we found that some plans were not fully executable due to motion-level planning failures, where the plans were not found in reasonable time.
Despite this phenomenon, our approach produces the most plan completions in all task settings on average (Figure~\ref{fig:enter-label}).
Although OLP was not always successful in execution, our approach generates plans that exhibit the highest success rates, matching the intention of the given instruction.
Interestingly, the \textit{spelling} task showed the lowest success rate in all approaches.
We attribute this to incorrect reasoning performed by the LLM at both the object and task levels, where the LLM may generate a plan sketch to stack the blocks in an incorrect or reversed order.

Although LLM-Planner generates plans without a solver, it does not complete a majority of tasks because the LLM poorly understands the configuration of the robot's environment for collision-free motion.
As a result, it incorrectly proposes actions that attempt to pick up an object blocked by another object or place an object in an occupied spot.
LLM+P also exhibits poor performance: although the LLM is capable of directly outputting PDDL, failures were mainly attributed to inaccurate problem definitions.
This may be due to the fact that LLM+P uses fewer prompts than OLP and DELTA; also, unlike DELTA, LLM+P does not provide definitions of PDDL planning operators, thus providing less context to the LLM.
We also observed that the PDDL problems generated by LLM+P and DELTA were susceptible to incorrect syntax, which is a drawback of LLM-based PDDL generation.
DELTA, whose approach closely resembles our method, performs better than LLM+P and LLM-Planner baselines, but it does not perform as well as our method while also generally requiring more tokens on average to generate planning definitions.
Similar to DELTA, OLP also demonstrates the advantage of bootstrapping task-level planning with PDDL subgoal definitions (reflected by low average planning times) but without relying upon the LLM to correctly generate PDDL definitions.
Our approach also requires less interaction with the LLM than DELTA as reflected by the average number of tokens.

\textbf{Limitations:} Much like how we humans plan, object-level planning serves as a critical interface between language and TAMP.
Our approach requires robot skill definitions specified as PDDL, which may not always transfer across robot systems.
However, we assume a set of FOON samples for few-shot learning. 
Further, we did not consider plan recovery if objects were knocked down during execution, thus lowering the success rate of completely executed plans.
Most importantly, as with baselines, this approach depends on a correctly generated object-level plan compatible with task-level planning for subgoal definitions.
In addition, our evaluations are performed solely on pick-and-place tasks, which do not highlight the benefits of the semantic richness of object-level plans.
In future work, we will explore broader task diversity and examine how we can use an LLM to adapt existing object-level plans to novel scenarios similar to prior work~\cite{paulius2018functional,sakib2022approximate}.
Like recent work~\cite{han2024interpret}, we can also integrate human feedback to correct LLM-generated errors at the object level.
We will also explore learning from demonstration to acquire task-level domain definitions to address our assumption of predefined skills.




\section{Conclusion}
\label{sec:conclusion}

We introduce a hierarchical planning approach that capitalizes on the power of large language models (LLMs) to bootstrap task and motion planning (TAMP).
Through an added layer of planning situated above TAMP known as object-level planning~\cite{paulius2022object}, we enable robots to flexibly find planning solutions from plan sketches extracted via LLM prompting.
Compared to alternative LLM-based planning approaches that either use an LLM as a planner or as a generator of planning definitions like PDDL~\cite{mcdermott1998pddl}, our method flexibly enables a robot to solve a wide range of tasks that greatly benefit from the expressiveness of natural language.


\section*{Acknowledgements}
This work was supported by the Office of Naval Research (ONR) under the REPRISM MURI N000142412603, ONR grants N00014-21-1-2584 and N00014-22-1-2592, Echo Labs, and the Austrian Science Fund (FWF) Project P36965 [DOI: 10.55776/P36965].
Partial funding for this work was provided by The Robotics and AI Institute (formerly ``The AI Institute").



\bibliographystyle{IEEEtrann}

\footnotesize{
\bibliography{refs}  
}

\onecolumn

\newpage

\appendices
\section{Object-level Planning Prompts}
\noindent

\begin{figure*}[h!]
\begin{lstlisting}[framexleftmargin=1mm,framextopmargin=1mm, framexrightmargin=1mm,framexbottommargin=1mm,mathescape]
$\textbf{System}$: You are a helpful assistant that will generate plans for robots. You will be given the following:
        1. A simple plan sketch, with which you will generate an entirely new plan sketch describing object states before 
            (preconditions) and after (effects) actions are executed.
        2. A list of objects available to the robot.

        Note the following rules:
        - Closely follow the task prompt. You must use all objects except any objects not related to the task.
        - Be consistent with object names throughout the plan.
        - All objects are on the table in front of the robot.
        - Use one action verb per step. However, any steps involving "pick" or "place" must be written as a single step with 
            the action "pick and place".
        - Use as many states as possible to describe object preconditions and effects.
        - Only use the states "in", "on", "under", or "contains" for describing objects. List them in the format 
            "$<$relation$>$ $<$obj$>$", where $<$relation$>$ is a state and $<$obj$>$ is a single object.
\end{lstlisting}
\caption{System prompt for our OLP method (refer to Section~\ref{sec:llm}).}
\label{fig:olp-sys-prompt}
\end{figure*}

\begin{figure*}[h!]
\begin{lstlisting}[framexleftmargin=1mm,framextopmargin=1mm, framexrightmargin=1mm,framexbottommargin=1mm,mathescape]
$\textbf{User}$: Your task will be to create a step-by-step plan for the following prompt: $\color{Bittersweet}\mathcal{T}$. The following objects are available in 
        the scene: $\color{Bittersweet}<\text{objects}\_\text{in}\_\text{scene}>$. Say 'Okay!' if you understand the task.

${\color{MidnightBlue}\text{\textbf{LLM}:~~Okay!}}$

$\textbf{User}$: Below are a list of prototype recipes. You must select the closest one that is the closest to the given task 
        prompt. Simply provide the number corresponding to the closest prototype.

        $\color{Bittersweet}<\text{Step-by-step language plans for examples }\mathcal{X}_{\mathcal{G}}=\{\mathcal{G}_1, \mathcal{G}_2,...,\mathcal{G}_n\}>$

${\color{MidnightBlue}\text{\textbf{LLM}:}<\text{Prototype selection } \hat{\mathcal{G}} \in \mathcal{X}_{\mathcal{G}}>}$

$\textbf{User}$: Generate a concise plan using the prototype as inspiration for the task: $\color{Bittersweet}\mathcal{T}$. Follow all guidelines. Give evidence to 
        support your plan logic.

${\color{MidnightBlue}\text{\textbf{LLM}:} <\text{Step-by-step instructions for task }\mathcal{T} \text{ as } \mathcal{P}_{L}>}$

$\textbf{User}$: Make a Python list of used objects in the following format: ["object$\_$1", "object$\_$2", ...]'. If there are several 
        instances of an object type, list them individually (e.g., ['first apple', 'second apple'] if two apples are used). Do not add any explanation.

${\color{MidnightBlue}\text{LLM:} <\text{List of objects needed for task }\mathcal{T}>}$

$\textbf{User}$: Format your generated plan as a JSON dictionary. List as many states as possible when describing each object's 
        preconditions and effects. Each required object should match a key in "object$\_$states": Be consistent with object names across actions. Use this JSON prototype as reference:

        $\color{Bittersweet}<\text{JSON equivalent of } \hat{\mathcal{G}}>$

${\color{MidnightBlue}\text{\textbf{LLM}:} <\text{JSON for task }\mathcal{T} \text{ as } \tilde{\mathcal{G}}_{\mathcal{T}}>}$
\end{lstlisting}
\caption{Progressive chain-of-thought (CoT) prompting for extracting object-level plans from a LLM (refer to Section~\ref{sec:llm}). We highlight special input provided to the LLM (particularly the user task query $\mathcal{T}$, objects in the scene, and few-shot examples---both as language instructions and JSON) as well as output acquired from the LLM in blue and orange colours respectively. A few-shot sample codified as a JSON structure is shown as Figure~\ref{fig:llm-a}.}
\label{fig:llm-planner-prompt}
\end{figure*}

\begin{figure*}[h!]
\begin{lstlisting}[framexleftmargin=1mm,framextopmargin=1mm, framexrightmargin=1mm,framexbottommargin=1mm,mathescape]
$\text{\{}$
    "plan": [$\text{\{}$
        "step": 1, 
        "action": "pick and place", 
        "required$\_$objects": ["first block", "second block"], 
        "object$\_$states": $\text{\{}$
            "first block": $\text{\{}$"preconditions": ["under nothing", "on table"], "effects": ["under second block", "on table"]$\text{\}}$, 
            "second block": $\text{\{}$"preconditions": ["under nothing", "on table"], "effects": ["on first block", "under nothing"]$\text{\}}$,
        $\text{\}}$, 
        "instruction": "Pick and place second block from table on first block."
    $\text{\}}$]
$\text{\}}$
\end{lstlisting}
\caption{JSON equivalent of a functional unit presented as Figure~\ref{fig:llm-to-olp}.}
\label{fig:llm-a}
\end{figure*}

\newpage

\section{Baseline Method Prompts}
\subsection{LLM-Planner}
\begin{figure*}[h!]
\begin{lstlisting}[framexleftmargin=1mm,framextopmargin=1mm, framexrightmargin=1mm,framexbottommargin=1mm,mathescape]
$\textbf{System}$: You are a helpful PDDL planning expert. Your job is to process a task prompt, a list of objects in the scene, and 
            a list of statements describing the environment state, reason about how to solve the task, and produce a plan that solves the task.

A task plan has the format of:
    1. ($<\text{action}\_\text{1}> <\text{arg1}> <\text{arg2}>$)
    2. ($<\text{action}\_\text{2}> <\text{arg1}> <\text{arg2}>$)
    3. ...

Observe the following rules:
- In the task plan, you can only use these actions:
    1. ($<\text{pick}> <\text{obj1}> <\text{obj2}>$) - pick $<\text{obj1}>$ that is on top of $<\text{obj2}>$; this causes nothing to be on $<\text{obj2}>$.
    2. ($<\text{place}> <\text{obj1}> <\text{obj2}>$) - place $<\text{obj1}>$ on top of $<\text{obj2}>$; $<\text{obj2}>$ must have nothing on it for $<\text{obj1}>$ to 
        be placed on it.
- Note the order of the arguments for both actions!
- The agent executing this task has a single hand: in order to pick up an object, the agent's hand must be free.

$\textbf{User}$: There is a scenario with the following objects: $\color{Bittersweet}<\text{objects}\_\text{in}\_\text{scene}>$. Please await further instructions.

$\textbf{User}$: Your task is as follows: $\color{Bittersweet}\mathcal{T}$. Transform this instruction into a PDDL goal specification in terms of 'on' relations. 
        Do not add any explanation.

${\color{MidnightBlue}\text{\textbf{LLM}:} <\text{Goal state description for task }\mathcal{T}>}$


$\textbf{User}$:: Find a task plan in PDDL to achieve this goal given the initial state below. Only specify the list of actions 
        needed.  Use the actions defined above. Do not add any explanation.

    Initial state: $\color{Bittersweet}<\text{State description as text }\tilde{s}>$

${\color{MidnightBlue}\text{\textbf{LLM}:} <\text{Task plan as text } \mathcal{P}_{\mu}>}$
\end{lstlisting}
\caption{Prompts for LLM-Planner baseline (Section~\ref{sec:baseline}). This baseline simply provides a textual description of the robot's actions as well as the current object configuration, with which it must reason about the correct sequence of actions that will result in collision-free execution and resolve the task.}
\label{fig:olp-prog-prompt}
\end{figure*}
\subsection{LLM+P}
\begin{figure*}[h!]
\begin{lstlisting}[framexleftmargin=1mm,framextopmargin=1mm, framexrightmargin=1mm,framexbottommargin=1mm,mathescape]

$\textbf{User}$: I want you to generate a PDDL problem file for robot problem solving. An example planning problem is:

        $\color{Bittersweet}<\text{PDDL Problem File Example}\tilde{s}>$

        Now I have a new planning problem and its description is as follows: These objects are on the table: 
        $\color{Bittersweet}<\text{objects}\_\text{in}\_\text{scene}>$. The current state of the world is: $\color{Bittersweet}<\text{State description as text }\tilde{s}>$.

        Your goal is to achieve this task: $\color{Bittersweet}\mathcal{T}$. Provide me with the problem PDDL file that describes the new planning problem  
        directly without further explanations.

${\color{MidnightBlue}\text{\textbf{LLM}:} <\text{PDDL Problem Definition}>}$
\end{lstlisting}
\caption[Prompts for the LLM+P~\cite{liu2023llm} baseline (Section~\ref{sec:baseline}). Similar to our OLP method, we provide the LLM with an example PDDL problem file definition for the most similar few-shot example.]{Prompts for the LLM+P~\cite{liu2023llm} baseline (Section~\ref{sec:baseline}). Similar to our OLP method, we provide the LLM with an example PDDL problem file definition for the most similar few-shot example. These prompts were based on those provided by \citet{liu2023llm} online.\footnotemark}
\label{fig:llm+p-prompt}
\end{figure*}

\footnotetext{LLM+P Repository:~\texttt{\url{https://github.com/Cranial-XIX/llm-pddl}}}

\newpage

\subsection{DELTA}
\begin{figure*}[h!]
\begin{lstlisting}[framexleftmargin=1mm,framextopmargin=1mm, framexrightmargin=1mm,framexbottommargin=1mm,mathescape]
$\textbf{User}$: Role: You are an excellent PDDL domain file generator. Given a description of action knowledge in natural language, 
        you can generate a PDDL domain file.

        Example: $\color{Bittersweet}<\text{PDDL Domain File Example}>$

        Instruction: A new domain includes the following objects: $\color{Bittersweet}<\text{objects}\_\text{in}\_\text{scene}>$. Please generate a corresponding new PDDL domain file for a robot. Do not add any explanation.

${\color{MidnightBlue}\text{\textbf{LLM}:} <\text{PDDL Domain Definition for task }\mathcal{T}>}$
\end{lstlisting}
\caption{PDDL domain file prompting for the DELTA~\cite{liu2024delta} baseline (Section~\ref{sec:baseline}). Similar to our OLP method, we provide the LLM with an example PDDL domain file definition for the most similar few-shot example.}
\label{fig:delta-domain-prompt}
\end{figure*}

\begin{figure*}[h!]
\begin{lstlisting}[framexleftmargin=1mm,framextopmargin=1mm, framexrightmargin=1mm,framexbottommargin=1mm,mathescape]
$\textbf{User}$: Role: You are an excellent PDDL problem file generator. Given a description of the robot's environment and a goal 
        description, you can generate a PDDL problem file.

        Example: $\color{Bittersweet}<\text{PDDL Problem File Example}>$

        Instruction: Now given a new description of the robot's scene and using the predicates in the previously generated PDDL domain file, please generate a new PDDL problem file for the task: $\color{Bittersweet}\mathcal{T}$. 

        $\color{Bittersweet}<\text{State description as text }\tilde{s}>$

${\color{MidnightBlue}\text{\textbf{LLM}:} <\text{PDDL Problem Definition for task }\mathcal{T}}>$

\end{lstlisting}
\caption{PDDL problem file prompting for the DELTA~\cite{liu2024delta} baseline (Section~\ref{sec:baseline}). Similar to our OLP method, we provide the LLM with an example PDDL problem file definition for the most similar few-shot example. This prompt was slightly modified to account for the lack of a scene graph in this work.}
\label{fig:delta-problem-prompt}
\end{figure*}

\begin{figure*}[h!]
\begin{lstlisting}[framexleftmargin=1mm,framextopmargin=1mm, framexrightmargin=1mm,framexbottommargin=1mm,mathescape]
$\textbf{User}$: Role: You are an excellent assistant in decomposing long-term goals. Given a PDDL problem file, you can decompose 
        the long-term goal in a sequence of subgoals.

        Example: $\color{Bittersweet}<\text{PDDL Subgoals Example}>$

        Instruction: Given the PDDL problem previously generated, please decompose the long-term goal into a sequence of subgoals considering the predicates and actions from the previously generated PDDL domain file. Simply list the decomposed PDDL subgoals for each instruction in a similar format as the example and only 1 level deep. 

${\color{MidnightBlue}\text{\textbf{LLM}:} <\text{PDDL Subgoals for task }\mathcal{T}>}$

\end{lstlisting}
\caption{Subgoal prompting for the DELTA~\cite{liu2024delta} baseline (Section~\ref{sec:baseline}). To accompany the few-shot example from the prior prompts, we also defined an example decomposition of subgoals as text. Our implementation of the baseline then extracts each set of subgoals to then generate PDDL subgoal problem files using the current state at the start of subgoal execution.}
\label{fig:delta-subgoal-prompt}
\end{figure*}

\newpage

\section{Task Settings}

\begin{figure}[h!]
    \centering
    \includegraphics[width=0.85\linewidth]{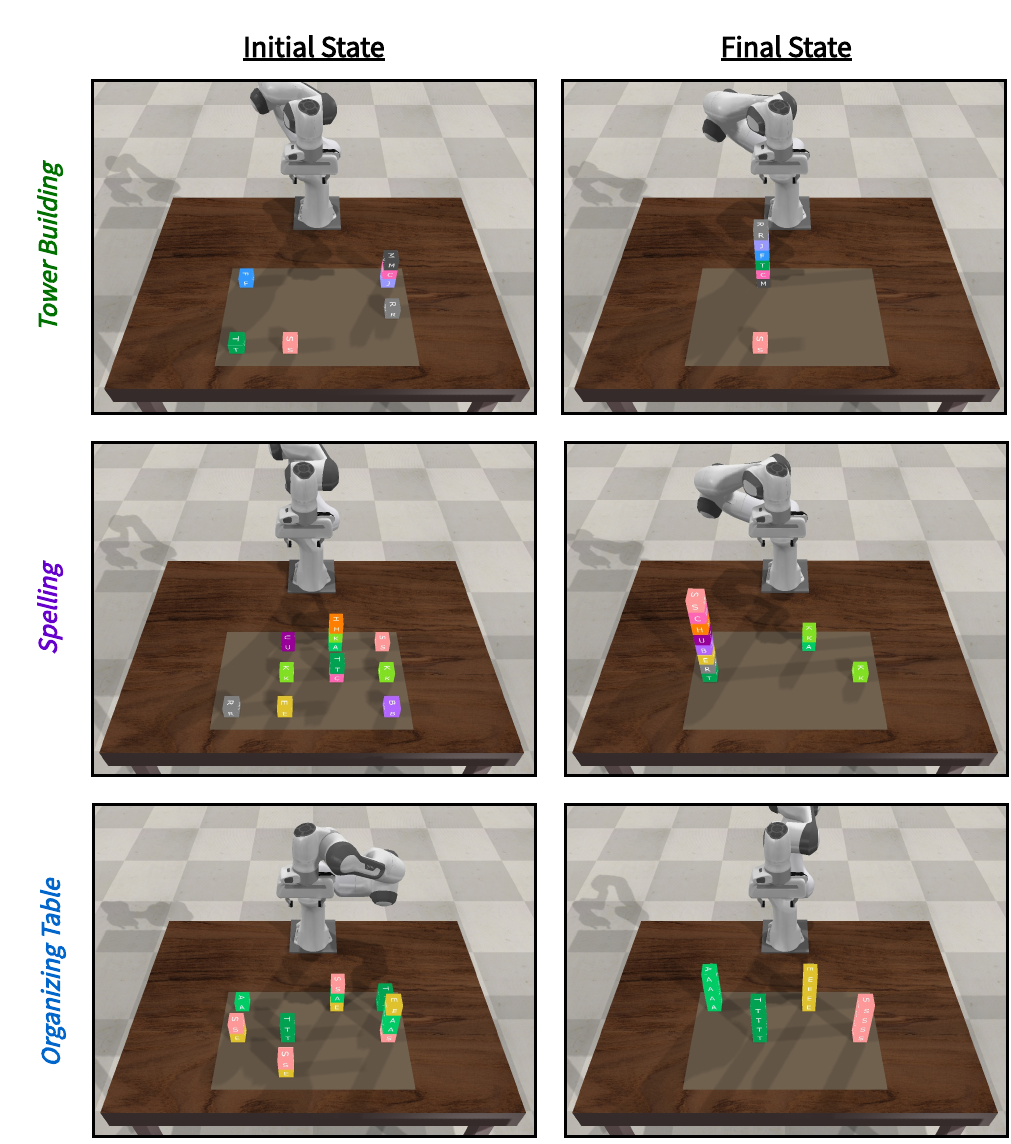}
    \caption{Examples of the initial and final states for all task settings defined in Section~\ref{sec:exp-setup}. In the \textit{tower building} example, the robot must simply construct a tower of 6 blocks (using all but one block on the table). In the \textit{spelling} task, the robot must assemble a tower spelling the word ``SCHUBERT" (read from top to bottom); this scene also contains extra blocks as distractions. Finally, in the \textit{organizing table} task, the robot must stack all alike blocks into separate towers or piles. There are a total of sixteen (16) blocks in this scene: four (4) block types, each with four (4) instances initially configured in a randomized order. In all settings, the object-level plan (OLP) will solely focus on the object types relevant to the task, as the LLM is not provided with the entire set of objects at the OLP generation phase (Section~\ref{sec:llm}).}
    \label{fig:exp-settings-full}
\end{figure}

\end{document}